\documentclass[journal]{IEEEtran}

\usepackage{xcolor,soul,framed} %,caption

\colorlet{shadecolor}{yellow}
\usepackage[pdftex]{graphicx}
\graphicspath{{../pdf/}{../jpeg/}}
\DeclareGraphicsExtensions{.pdf,.jpeg,.png}
\usepackage{graphicx}
\usepackage[cmex10]{amsmath}
\usepackage{array}
\usepackage{mdwmath, stmaryrd, amssymb}
\usepackage{mdwtab}
\usepackage{eqparbox}
\usepackage{url}
\usepackage{subfig}
\usepackage[hidelinks]{hyperref}
\usepackage[colorinlistoftodos]{todonotes}
\newcounter{MYtempeqncnt}

\usepackage{academicons}
\definecolor{orcidlogocol}{HTML}{A6CE39}

\begin{document}

\bstctlcite{IEEEexample:BSTcontrol}
    %\title{\LARGE{Exercise Recommendation System with Real Time Active Learning based on Marginal Distance Probability Distribution}  }
    %\title{\LARGE{Real-Time Learning using an Expert with a Deep Recommendation System with a Marginal Distance Probability Distribution}  
    \title{\LARGE{Real-Time Learning from An Expert in Deep Recommendation Systems with Application to mHealth}
    }
  \author{Arash Mahyari \href{https://orcid.org/0000-0001-8660-3096}{\includegraphics[scale=0.1]{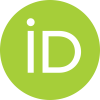}},  Peter Pirolli, Jacqueline A. LeBlanc \\
      
  %\thanks{Manuscript received July 10, 2012. This work was funded in part by NIH under Grant xxxxxxxxx.}
  \thanks{Research reported in this paper was supported by the National Institute on Aging  of the National Institutes of Health under award number R01AG053163.}
  \thanks{A. Mahyari is a Research Scientist at Florida Institute for Human and Machine Cognition (IHMC), Pensacola, FL 32502 USA (e-mail: amahyari@ihmc.org).}% <-this % stops a space
  \thanks{P. Pirolli is a Senior Research Scientist at Florida Institute for Human and Machine Cognition (IHMC), Pensacola, FL 32502 USA (e-mail: ppirolli@ihmc.org).}%
  \thanks{J. LeBlanc is a certified health and fitness coach and independent consultant.}%
  \thanks{The code repository: \url{https://github.com/arashmahyari/ExerRecomActiveLearn}.}
  }

% The paper headers
%\markboth{IEEE Journal of Biomedical and Health Informatics, VOL.~xx, NO.~xx, 2021
\markboth{IEEE Journal of Biomedical and Health Informatics, 2022
} {Mahyari \MakeLowercase{\textit{et al.}}: Real-Time Learning from An Expert in Deep Recommendation Systems with Application to mHealth}

% ====================================================================
\maketitle

%\todo[inline]{Need a better title}

% === ABSTRACT ====================================================================
% =================================================================================
\begin{abstract}
%\boldmath
Recommendation systems play an important role in today's digital world. They have found applications in various areas such as music platforms, e.g., Spotify, and movie streaming services, e.g., Netflix. Less research effort has been devoted to physical exercise recommendation systems. Sedentary lifestyles have become the major driver of several diseases as well as healthcare costs. In this paper, we develop a recommendation system to recommend daily exercise activities to users based on their history, profiles and similar users. The developed recommendation system uses a deep recurrent neural network with user-profile attention and temporal attention mechanisms. 

Moreover, exercise recommendation systems are significantly different from streaming recommendation systems in that we are not able to collect click feedback from the participants in exercise recommendation systems. Thus, we propose a real-time, expert-in-the-loop active learning procedure. The active learner calculates the uncertainty of the recommendation system at each time step for each user and asks an expert for recommendation when the certainty is low. In this paper, we derive the probability distribution function of \textit{marginal distance}, and use it to determine when to ask experts for feedback. Our experimental results on a mHealth and MovieLens datasets show improved accuracy after incorporating the real-time active learner with the recommendation system.

%The implementation can be found at: \url{https://github.com/arashmahyari/PolicyAugmentation}.

\end{abstract}

% === KEYWORDS ====================================================================
% =================================================================================

% For peer review papers, you can put extra information on the cover
% page as needed:
% \ifCLASSOPTIONpeerreview
% \begin{center} \bfseries EDICS Category: 3-BBND \end{center}
% \fi
%
% For peerreview papers, this IEEEtran command inserts a page break and
% creates the second title. It will be ignored for other modes.
\IEEEpeerreviewmaketitle

% ====================================================================
% ====================================================================
% ====================================================================

% === I. INTRODUCTION =============================================================
% =================================================================================
\section{Introduction}

%\IEEEPARstart{A}{he}    

A major driver of healthcare costs in different countries are unhealthy behaviors such as physical inactivity, increased food intake, and unhealthy food choice \cite{riley2015news, thorpe2009future}. Behavioral and environmental health factors account for more deaths than genetics \cite{riley2015news}. Pervasive computational, sensing, and communication technology can be leveraged to support individuals in their everyday lives to develop healthier lifestyles. For instance, the pervasive use of smartphones is a potential platform for the delivery of behavior-change methods at great economies of scale. Commercial systems such as noom \cite{noom} aim to provide psychological support via mobile health (mHealth) systems. Research platforms, such as the Fittle+ system \cite{pirolli2018scaffolding}, have demonstrated the efficacy of translating known behavior-change techniques \cite{michie2014abc} into personal mHealth applications. However, most work in the mHealth domain are limited to the development of smartphone applications and connecting subjects and expert knowledge.

On the other hand, recommendation systems are becoming popular in various applications. E-commerce websites have been using recommendation systems to suggest new products and items to existing and new users to entice them into purchasing new items \cite{linden2003amazon, bodapati2008recommendation}. With the advent of streaming movies and music services, e.g., Netflix, Spotify, service providers need to keep their users interested or lose their customers and profits. Thus, these streaming service providers deploy recommendation systems to suggest new movies and musics to their existing and new users based on their history of watching movies and listening to musics \cite{bennett2007netflix, song2012survey}. However, not many research studies have been devoted to developing recommendation systems for exercise activities.

In this paper, we develop an attention-based recommendation system for exercise activities to new users using a mHealth application. The proposed recommendation system is based on a deep recurrent neural network that takes advantage of users' profiles and exercise characteristics as features and temporal attention mechanisms. However, one major difference between exercise activities and other domains is that the recommendation system is not able to collect users' feedback. In movie, e-commerce, and music applications, recommendation systems constantly receive feedback from users' clicks data. When a recommendation system suggests a new music (or a movie) to a user, the user may click on the suggested music (or movie) and listen to (or watch) it completely. The click and duration of playback is used as a feedback to fine-tune the recommendation system. Although users are able to provide ratings in these systems, they may choose not to do so. As a result, the systems will use the click data, duration of playback, etc. to assess whether the recommended music (or movie) was a good recommendation or not.

However, when a recommendation system suggests an exercise activity to the user, the application is not able to collect the click data because the application cannot observe the user, and the only way to collect the feedback is if the user chooses to rate the exercise. In other words, the recommendation system doesn't know whether the user completed the exercise, i.e. if it was a good recommendation. In some mHealth apps \cite{konrad2015finding}, users were asked to provide that information manually, but the missing data is tremendous as most users were ignorant of that feedback. The problem is more challenging when the recommendation system faces a new user without any history. To address this issue, in this paper, we take advantage of a real time expert-in-the-loop mechanism. Individuals trust expert personal trainers to provide them with exercises plans based on the experts' knowledge and experience. Our proposed system will leverage this trust to use experts' knowledge when the system is uncertain. As a result, more individuals will have access to exercises plans with expert personal trainers in the loop. Our proposed network will calculate the certainty of a new recommendation using the probability distribution of the \textit{marginal distance}, the difference between the highest probability and the second highest probability of exercise classes in the output of the recommendation system. To quantify the certainty, we derive the probability distribution of the marginal distance from the probability distribution of the last layer of the recommender. Even though the marginal distance has been used in active learning, this is the first work to provide its probability distribution for statistical hypothesis testing.

\begin{figure}[t]
    \centering
    \includegraphics[width=0.97\columnwidth]{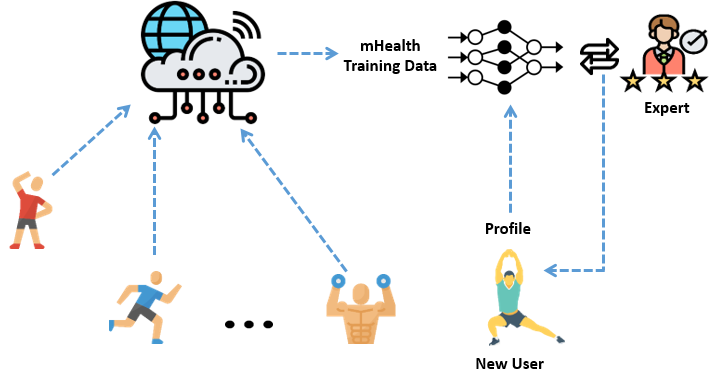}
    \caption{The overall architecture of the proposed recommendation system with expert-in-the-loop. The exercise activities are recommended through a smartphone app, and their completion is collected from smartphones. The deep recommendation system is trained on the collected history data and its augmentation. The new recommender system is initialized for each new participant from the global trained model, and fine-tuned with similar users based on their profiles. At each time step, a new exercise is recommended to users. If the recommendation system is uncertain about the new recommendation, \textit{i.e.} whether the user will complete the exercises or not, the recommendation system will ask the expert for correction.}
    \label{fig:overal}
    \vspace{-5mm}
\end{figure}

The other challenge with most recommendation systems is the initialization of the recommendation system for new users. Much work has been devoted to address this issue, including metal-learning approaches \cite{bharadhwaj2019meta}. In this paper, we leverage the questionnaire filled by users and their demographic information to find the existing users with similar interests and demographic information. Then, the global recommendation model is fined tuned with the history of the similar users. Comparing to other existing methods, this approach has less complexity. Fig.~\ref{fig:overal} shows the overall architecture of the recommendation system.

The rest of this paper is organized as follows: Section~\ref{sec:data} describes the mHealth data used in this study. The architecture of the proposed recommendation system is explained in Section~\ref{sec:network-arch}. Section~\ref{sec:new-user} explains how we take advantage of the user profiles to initialize a recommendation system for new users. Section~\ref{sec:active-learning} describes the proposed active learning procedure based on the distribution of the \textit{marginal distance}. Section~\ref{sec:experimental} is devoted to the experimental results and discussion.

\subsection{Related Work}

More recently, mHealth systems have found applications in different healthcare domains since the advent of smart phones. Dunsmuir \textit{et al.} \cite{dunsmuir2014development} developed mHealth for diagnosis and management of pregnant women with pre-eclampsia. In \cite{seepers2015enhancing}, the inter-pulse-interval security keys was used to authenticate entities for various mHealth applications. Schiza \textit{et al.} \cite{schiza2018proposal} proposed a unified framework for an eHealth national healthcare system for European Union. In another work \cite{huang2013we}, WE-CARE, a mobile $7$-lead ECG device, was developed to provide 24/7 cardiovascular monitoring system. In \cite{wuttidittachotti2015mhealth}, an mHealth system was designed and developed to provide exercise advice to participants based on their Body Mass Index, Basal Metabolic Rate, and the energy used in each activity or sport, e.g. aerobic dancing, cycling, jogging working and swimming. However, this work does not use machine learning algorithms.

On the other hand, recommendation systems have been used in e-commerce and online shopping for several years \cite{reddy2019content, khoali2020advanced, panagiotakis2021improving, fessahaye2019t}. The goal of recommendation systems is to recommend products that suit the consumers’ tastes. Traditional recommendation systems have used collaborative filters to suggest products similar to those the consumers have purchased. With the advancements in deep learning algorithms \cite{ravi2016deep}, several studies have proposed deep learning-based recommendation systems \cite{liu2016recurrent}. In \cite{liu2016recurrent}, a multi-stack recurrent neural network (RNN) architecture is used to develop a recommendation system to suggest businesses in Yelp based on their reviews. Wu \textit{et al.} \cite{wu2017recurrent} used an RNN with long-short term memories (LSTM) to predict future behavioral trajectories. A few studies have proposed exercise recommendation systems \cite{sami2008design}. Sami \textit{et al.} \cite{sami2008design} used several independent variables to recommend various sports such as swimming using collaborative approaches. Ni \textit{et al.} \cite{ni2019modeling} developed an LSTM-based model called FitRec for estimating a user’s heart rate profile over candidate activities and then predicting activities on that basis. The model was tested against 250 thousand workout records with associated sensor measurements including heart rate.

In the health care domain, Yoon \textit{et al.} \cite{yoon2016discovery} developed a recommendation system for a personalized clinical decision making. The system used electronic health records of different patients and their clinical decisions to recommend clinical decisions for new patients. In a recent work \cite{liu2020dna}, support vector machines (SVMs), random forest, and logistic regression were used to recommend skin-health products based on genetic phenotypes of consumers. In a similar study \cite{li2018adaptive}, user contextual features and daily trajectories of steps over time were used to develop a recommendation system for planning an hour-by-hour activity. The model classifies users to subgroups and recommend activities based on the history of similar users. The proposed method doesn't use any time series modeling, e.g., recurrent neural networks, to learn patterns, thus lacking the ability to generalize to more users. 

In a related field, several works have been devoted to the cold-start problem in recommendation systems \cite{lee2019melu, li2019zero}. The cold-start problem refers to the new users whose historical data is not available to the recommendation system. Thus, the recommendation system is not able to accurately recommend items, e.g., music, movies, etc., to the new users. Meta-learning approaches try to learn a global model from all users to initialize the recommendation systems for new users \cite{lee2019melu}. The shortcoming of these approaches is that they are not 
personalized in the beginning. The second approach is zero-shot, one-shot, or transfer learning methods \cite{li2019zero, mahyari2021robust}. These approaches transfer a global model learned from other users' historic data to new users. These approaches have great performance in classifiers, but they still have shortcomings as recommendation systems dealing with sequences of data. The proposed approach leverage experts (human personal trainers) to actively learn personalized exercise programs for new users. While many people use trainers to get recommendation for their daily exercise activities, the proposed approach leverages the expert knowledge to reach a broader group of users at a lower cost because the expert does not need to be continuously involved in recommending new exercises, just in the beginning when the system is uncertain about new users.

\section{\lowercase{m}Health Data}
\label{sec:data}

The data we use comes from the Konrad et al. \cite{konrad2015finding} mHealth experiment with DStress. It was developed to provide coaching on exercise and meditation goals for adults seeking to reduce stress.  The purpose of the experiment was to test the efficacy of an adaptive daily exercise recommender (DStress-adaptive) against two alternative exercises programs in which the daily exercises changed according to fixed schedules (Easy-fixed and Difficult-fixed). The DStress-adpative recommender was a hand engineered finite state machine. The transition rules are described in more detail in Konrad et al. \cite{konrad2015finding}, but they implement a policy whereby, if a person successfully completes all three exercises assigned for a day, they advance to the next higher level of exercise difficulty. If they do not succeed at exercises or meditation activities, then they are regressed to exercises or meditation activities at an easier level of difficulty.  The 44 exercises used in DStree and their difficulty ratings were obtained from three certified personal trainers (e.g., Wall Pushups, Standing Knee Lifts, Squats, and Burpees, etc.). The experiment took place over a 28-day period. In a given week, users encountered three kinds of days: Exercise Days (occurring on Mondays, Wednesdays, Fridays), Meditation Days (Tuesdays, Thursdays, Saturdays), and Rest Days (Sundays).  

72 adult participants (19-59 yr) were randomly assigned to three conditions with different 28-day goal progressions: (1) a DStress-adaptive condition using the adaptive coaching system in which goal difficulties adjusted to the user based on past performance, (2) an Easy-fixed condition in which the difficulty of daily goals increased at the same slow rate for all participants assigned to that condition, and (3) a Difficult-fixed condition in which the goal difficulties increased at a greater rate. Konrad et al. \cite{konrad2015finding} found that the adaptive DStress-adaptive condition produced significant reductions in self-reported stress levels compared to the Easy-fixed and Difficult-fixed goal schedules.  The DStress-adaptive condition also produced superior rates of performing assigned daily exercise goals.

\noindent \textbf{User Profile:} A variety of pretest survey data were collected in the Konrad et al. \cite{konrad2015finding} study that provides our user data. These included the (1) Perceived Stress Scale (PSS), which is a 10-item psychometric scale assessing perceived stress over the past month, (2) Depression, Anxiety, Stress Scale (DASS), a 21 item assessment of depression, anxiety, and stress, (3) the Cohen-Hoberman Inventory of Physical Symptoms (CHIPS): a 33-item scale measuring concerning physical symptoms over the past 2 week, (4) BMI, the Body Mass Index, and (5) Goldin Leisure-Time Exercise Questionnaire (GLTEQ), a 4-item scale measuring frequency of physical activity during leisure time, and (6) the Exercise Self-Efficacy Scale (EXSE), and 8-item assessment of self-efficacy about exercising in next 1-8 weeks. 

The Perceived Stress Scale (PSS) is the most widely used instrument for measuring the perception of street \cite{cohen1983global, cohen1988perceived}. The Depression Anxiety Stress Scale \cite{coker2018psychometric} is a valid, reliable instrument measuring depression, anxiety, and stress \cite{antony1998psychometric}. The Cophen-Hoberman Inventory of Phsyical Symptoms (CHIPS) measures concerns about physical symptoms over the past 2 weeks was included because stress is often manifested in such symptoms \cite{mcfarlane1994physical}. The Godin Leisure-Time Exercise Questionnaire (GLTEQ; 4 item test) was included to assess pre-experimental activity levels \cite{godin2011godin}. The Exercise Self-efficacy Scale (EXSE) is an 8-item test assessing individuals’ beliefs in the ability to exercise \cite{mcauley1993self}.

\noindent \textbf{Exercise Profile:} The original Konrad et al. study \cite{konrad2015finding} obtained difficulty ratings of the exercises from three subject matter experts (SMEs; personal trainers) that were predictive of the probability of performing the exercises \cite{pirolli2016computational}. We augmented these data with exercise classification, attributes, and relations obtained from a 60 minute structured interview with an SME (fitness coach) that was followed up with specific clarification questions. The structured interview consisted of a card-sorting task and an exercise program planning task.

For the card-sorting task, exercise names and descriptions were placed on 3x5 cards. The cards were shuffled and the SME was asked to go through the cards to familiarize herself with the exercises. The SME was asked to sort the exercises into piles by whatever criteria “seemed natural.” The SME was asked to label those piles. These original piles were grouped into super-categories and labeled. Then the original piles were sorted into subcategories and labeled recursively until no further subgroups made sense to the SME. The SME was then asked if there was a possible “alternative grouping” of the exercises. The SME was also asked to rate the difficulty of each exercise. This card sorting produced an initial hierarchical classification of the exercises into a top level of resistance exercises and metabolic conditioning exercises. Within those supercategories there were subcategories for push, pull, squats, lunges, single leg stance, and core exercises. Further subcategories consisted of back, chest, legs. legs/glutes, core/abs, and abs/glute exercises. The alternative grouping consisted of full body, compound, and power categories.

% The exercise program planning tasks involved asking the SME to produce program plans for client personas with high/low motivation or high/low experience and skill. They were asked to generate multi-week programs consisting of sequences of exercises (including number of sets and reps) for each day, and how those sequences would change ideally over the days and weeks of the program. Specific questions were asked about the goals of the programs for the end-user, how that client was expected to change by the end of the program, possible issue and barriers, and how to respond to each barrier/issue with changes in exercises.

\section{Proposed Recommendation System}

The goal of this paper is to recommend the next exercise for a given participants based on the history of exercises the participant completed. Let $X^i(0), X^i(1), \ldots, X^i(T)$ represent the exercise history for the $i$th participant, where $X^i(.) \in \mathbb{R}^{N \times 1}$ is the one-hot encoding of the exercise and $N$ is the total number of exercises ($N=44$ in mHealth dataset). Let $U^i \in \mathbb{R}^{N_U \times 1}$ and $E^j \in \mathbb{R}^{N_I \times 1}$ represent the $i$th participant's profile and the $j$th exercise's profile, respectively. The recommender gets $X^i(t-\omega+1), \ldots, X^i(t-1)$ as the input in addition to the $i$th user's profile and the exercises' profiles and predicts $X^i(t)$ to recommend to the user. The length of the window, $\omega$, is determined by the developer. In this paper, we use autocorrelation function (ACF) of $X$ to determine the appropriate window length, similar to the statistical time series analysis methods \cite{hamilton2020time}.

\subsection{Network Architecture} 
\label{sec:network-arch}

The proposed network consists of five modules: encoder, decoder, recurrent neural network (RNN), user attention, exercise temporal attention. The hyper-parameters of this architecture are selected empirically. 

\noindent \textbf{Exercise input:} The encoder is a fully connected (FC) linear layer that embeds the $N$-dimensional exercise names onto a $K_X$-dimensional vector space:

\begin{equation}
    H^i_X(t)=ReLU(W_X \times X^i(t)),
\end{equation}

\noindent where $H^i_X(t) \in \mathbb{R}^{K_X \times 1}$ is the exercise name's embedding at time $t$ for the $i$th participant, $W_X \in \mathbb{R}^{K_X \times N}$ is a trainable weight matrix learned from the training data, and \textit{ReLU} is the activation function.

\noindent \textbf{User Profile input:} The profile of the $i$th user $U^i \in \mathbb{R}^{N_u \times 1}$ is provided as a $N_u$-dimensional vector. The user profile is embedded onto a $K_u$-dimensional vector space:

\begin{equation}
    H^i_u=ReLU(W_u \times U^i),
\end{equation}

\noindent where $H^i_u \in \mathbb{R}^{K_u \times 1}$ is the embedding of the $i$th user's profile, and $W_u \in \mathbb{R}^{K_u \times N_u}$. 

Users' profiles, e.g., demographic information,  provide valuable information about paying attention to specific aspects and features of exercises. For example, age and gender are two important variables that can significantly affect the types of exercises users will likely to perform. The attention mechanism will highlights the features of the exercise (extracted by neural network) that are more relevant to the current user based on their demographic information. Thus, we combine the exercise name embedding and the user profile embedding using:

\begin{equation}
    p_u(t)=softmax(W_1 \times H^i_x(t) \oplus W_2 \times H^i_u),
\end{equation}

\noindent where $W_1 \in \mathbb{R}^{K \times K_x}$ and $W_2 \in \mathbb{R}^{K \times K_u}$ are trainable mapping matrices, $p_u(t) \in \mathbb{R}^{K \times 1}$ is the combination vector of the exercise and the user profiles which provides the attention probability for each entity of the input exercise. The final vector is the element-wise multiplication of the exercise name embedding and the attention probability: $\psi^i(t)=p_u(t) \odot (W_1 \times H^i_x(t))$. The exercise name embedding $H^i_x$ is a vector in a $K_x$-dimensional space, and multiplying it with the attention probability $p_u(t)$ rotates this vector in the $K_x$-dimensional space.

\begin{figure}[t]
    \centering
    \includegraphics[width=0.97\columnwidth]{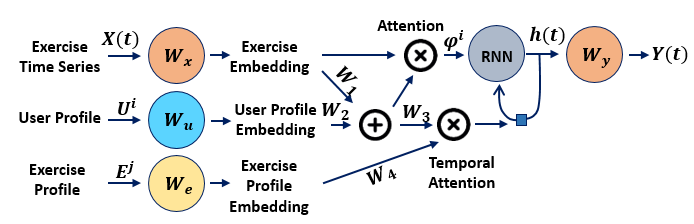}
    \caption{The architecture and modules of the proposed deep recurrent neural network. The recommendation system uses users and exercise profiles as attention mechanism. The attention mechanism will highlight the most relevant characters tics of the exercises to each user.}
    \label{fig:architecture}
\end{figure}

\noindent \textbf{Exercise Profile input and temporal attention mechanism:} Similarly, the profile of the $j$th exercises is provided with a $N_e$-dimensional vector $E^j \in \mathbb{R}^{N_e \times 1}$ and embedded using a linear layer:

\begin{equation}
    H^j_e(t)=ReLU(W_e \times E^j(t)),
\end{equation}

\noindent where $H^j_e(t) \in \mathbb{R}^{K_e \times 1}$ is the embedding of the $j$th exercise, and $W_e \in \mathbb{R}^{K_e \times N_e}$ is the trainable matrix. Note that we use $(t)$ in front of the exercise profile similar to the exercise names $X(t)$ to represent the time information.

At each time step, the user performs an exercise that has a huge impact on the future exercises the user has desire to complete. For example, an user who is doing exercises focused on upper body may want to continue upper-body exercises for a day and then focus on the lower-body exercises on another day. In another example, the difficulty of the current exercise affects the difficulty level of the short-term future exercises. To incorporate this information into the recommendation system, we use the exercise profiles as an attention mechanism to give weights to different exercises at different time steps. The temporal attention mechanism assigns probability values to different time steps within the $\omega$-length window. The exercise embeddings are combined with $\psi^i(t)$:

\begin{equation}
    p_e(t)=softmax(W_3 \times \psi^i(t) \oplus W_4 \times H^j_e(t)),
\end{equation}

where $W_3 \in \mathbb{R}^{\omega \times K}$ and $W_4 \in \mathbb{R}^{\omega \times K_e}$ are trainable mapping matrices, $p_e(t) \in \mathbb{R}^{\omega \times 1}$ is the combination vector of the user-attention exercise and exercise profiles giving the attention probability for each time step. The final time series of vectors with length $\omega$ used as the input to the RNN module is: $\phi^i(t-k)=p_e(k) \times \psi^i(t-k)$ for $k=\{1,2,\ldots, \omega \}$.

\noindent \textbf{RNN:} The RNN module is responsible for learning the sequential pattern of the exercise history. Although there are different variants of RNN modules with long short-term memory (LSTM) and gated recurrent units (GRU), we will use regular RNN modules. The reason is that LSTMs and GRUs have built-in units to learn the dependencies of time series over time and accentuate or de-emphasize relevant information at different time steps. In this paper, we will use temporal attention mechanism that will take into account the importance of different time steps. Moreover, our auto-correlation function (ACF) analysis of this dataset shows a short-term dependencies for exercise recommendation. However, in developing recommendation systems for datasets with long-term dependencies, we will replace regular RNNs with RNNs with LSTM or GRU units. The RNN module consists of one hidden layer with \textit{ReLU} activation function:

\begin{equation}
    h(t)=ReLU(W_{\phi} \times \phi^i(t)+W_h \times h(t-1)),
\end{equation}

\noindent where $W_{\phi}$ and $W_h$ are trainable weights and $h(t)$ is the hiddent state at time $t$.

\noindent \textbf{Exercise name prediction:} The decoder converts the predicted hidden state at time $t$ into exercise names: 

\begin{equation}
    Y^i(t)=softmax(W_y \times h(t)),
\end{equation}

\noindent where $W_y$ is a trainable matrix and $Y^i(t) \in \mathbb{R}^{N \times 1}$ is the multinomial distribution over $N$ exercises. 

\noindent \textbf{Training:} The whole network is trained end-to-end with the \textit{cross entropy} loss function:

\begin{equation}
    \min -\mathbb{E}_{X(t) \sim p_{data}}[log(P(Y(t)|X(t-1), \ldots, X(t-\omega), U, E)]
\end{equation}

All modules are analytically differential. Thus, the gradient can back-propagate through decoder, recurrent layers through time, and user and temporal attention modules. Adam optimizer is used for training the network \cite{kingma2014adam}.

\subsection{New User Initialization}
\label{sec:new-user}

The recommendation system is trained with a set of training data collected from different users. The system is general and not personalized for new users. To personalized the recommender, we fine tune our network with the training dataset of the users whose profiles are very similar to the new user. To achieve this goal, we calculate the similarity between $U^i$ and the existing users by Euclidean distance, $\sqrt{\|U^i-U^j\|^2}$ for $i \neq j$, to find the most similar users. Then, users are sorted from the most similar to the least similar, and the $k$ most similar users are selected. Afterwards, the recommender system is fine-tuned with the training data of the selected users, small learning rate, and one epoch.

\section{Active Learning}
\label{sec:active-learning}

Active learning is used to ask experts for providing annotation for unlabeled samples \cite{aghdam2019active, zhang2020state}. However, it is impossible to ask experts for annotating a large set of unlabeled samples. The active learner is usually presented with a limited budget to ask experts for annotation. The active learner selects the most informative samples based on the uncertainty that a trained classifier has about these samples. Two most common methods of measuring the uncertainty are entropy and \textit{marginal distance} \cite{wang2016cost, joshi2009multi, gu2019recursive}. 

In this paper, we use active learning to personalize the recommender for each user. The recommender is trained on a set of training data from different people. Thus, the recommender is not tailored for new users. The profile of the users will provide the attention mechanism in the feature space, but not enough to personalize the recommender. When we get a new user, the recommender will be used. However, when the recommender is uncertain about the next recommended exercise, the recommender will ask the expert to intervene and provide the next recommendation. We will use the marginal distance and entropy of the series of recommendation as criteria to decide when to ask the expert. However, the existing work has set an arbitrary threshold on the marginal distance. In this paper, we derive the probability distribution function of the marginal distance.

\subsection{Marginal Distance Random Variable}
\label{sec:marg-dist}

The output of the classifier (the last layer of the recommender) is a vector of $N$ random variables, $Y=[y_1, y_2, \ldots, y_N]^T$, with a Multinomial distribution. Ideally, one of the $y_i$s is one and the rest are zeros. However, $N$ random variables represent the probability of the input sample belonging to each of these classes. Let $\bf {p}=[p_1, p_2, \ldots, _N]^T$ represent the probability values. $p_i$ has a Beta distribution and the vector $\bf {p}$ is drawn from a Dirichlet distribution \cite{forbes2011statistical}. Thus, these $N$ random variables are sorted in ascending order $y_{(1)} \leq y_{(2)} \leq \ldots \leq y_{(N)}$ and the \textit{marginal distance} is defined as $Z=y_{(N)}-y_{(N-1)}$. The distribution of the \textit{marginal distance} is (see Appendix~\ref{sec:appendix} for proof): 

\begin{multline}\label{eq:active-learning1}
    f(z)= {  {\Gamma (\alpha_1+\alpha_2+\alpha_3)} \over {\Gamma(\alpha_1)\Gamma(\alpha_2)\Gamma(\alpha_3)}} \\
     \int_{y_{(N-1)}=0}^{1-z \over 2}
     {(y_{(N-1)}+z)}^{(\alpha_1-1)} y_{(N-1)}^{(\alpha_2-1)}\\ {(1-2y_{(N-1)}-z)}^{(\alpha_3-1)} dy_{(N-1)}
\end{multline}

The Monte Carlo method will be used to approximate this distribution.

\subsection{Active Learning Procedure}

The key component in active learning is to determine when to ask the expert for feedback. The recommender system needs to ask for the expert opinion when the input sample is out of the distribution of the training dataset. Let $\mathcal{D}_s$ represents the distribution of the training dataset. We use $\mathcal{D}_s$ to derive the probability distribution function as the \textit{marginal distance} as defined in \S~\ref{sec:marg-dist}. Let $\mathcal{D}_\mathcal{M}$ represent this distribution, and $z^i(t)$ represent the \textit{marginal distance} for the $i$th user at time $t$. Then, our hypothesis is:

\begin{equation*}
    \begin{cases}
      H_0: & z^i(t) \in \mathcal{D}_\mathcal{M}\\
      H_1: & \text{otherwise}
    \end{cases}
\end{equation*}

The $\alpha$-level hypothesis testing determines whether the recommended exercise should be given to the user or ask the expert for the right exercise recommendation (feedback or label). Then, the trained recommender is fine-tuned with the feedback from the expert for personalization.

\renewcommand{\thefootnote}{}
\section{Experimental Results} \footnote{The code repository: https://github.com/arashmahyari /ExerRecomActiveLearn.}
\label{sec:experimental}

In this section, the proposed approach with different components are evaluated on the offline mHealth dataset. The proposed architecture is also evaluated on the MovieLens100K dataset \cite{harper2015movielens}. 

\subsection{mHealth}

This section is devoted to the evaluation of the proposed exercises recommendation system on an exercises activity dataset. The dataset was described comprehensively in Section~\ref{sec:data}.

\subsubsection{\textbf{Data Augmentation}}
\label{sec:augment}

The challenge with recommendation systems is not having enough training data. One way to increase the amount of training data is to use data augmentation \cite{yu2019hierarchical, kafle2017data}. While data augmentation in computer vision is straightforward and can be achieved by adding noise or cropping images randomly, it requires careful attention for sequential and symbolic data, e.g., language \cite{yu2019hierarchical}. The random creation of sequential data has negative effect as it introduces noise to the data that does not follow the sequential pattern in the real data. 

For exercise activities, there are two ways to augment the sequential data: asking the human expert or using association mining on the training data. In the first approach, an exercise expert was asked to categorize different exercises available for participants. Then, the augmentation algorithm goes over the sequence of exercises for each participant in the training data, chooses $10\%$ of exercises and replace them with similar exercises in the same category. 

In the second approach, we propose to use association rule mining \cite{tan2016introduction} to extract rules from frequent itemsets. Then, the augmentation algorithms go over the sequence of exercises of each participant, choose $10\%$ of exercises and replace them with their similar exercises based on these rules.

\subsubsection{\textbf{Experiment Setup and Discussion}}

In order to find the appropriate length of window for the RNN model, we looked at the autocorrelation function of the exercise sequence. The autocorrelation function shows the degree of the dependencies of time series data and is often used to select the order of the time series analysis methods. Our ACF analysis shows that the length of the sequence for the RNN model should be $w=3$. Processing the data with $w=3$ results in $2343$ sequence of training samples. The architecture of the proposed recommendation system is selected empirically as follows: $K_X=20$, $K_u=15$, $K_e=3$, and $K=10$.

\vspace{2mm}

\noindent \textbf{Baseline:} We use the proposed RNN model with the user profile attention mechanism and the exercise profile temporal attention mechanism (described in $\S$~\ref{sec:network-arch}) as our baseline method. The model is trained with \textit{Cross-entropy} loss function and Adam optimizer \cite{kingma2014adam} for $30$ epochs. We use \textit{k-fold} approach for evaluating the performance of the recommender. In our \textit{k-fold} setup, we kept one participant out of the training dataset and used the remaining $71$ participants data for the training. Then, the left-out participant is treated as a new participant and the trained recommender is used to recommend exercises to the new participant. The actual data from the left-out participant is used as the groundtruth to calculate the accuracy of the recommendation system. We calculate the \textit{top-1}, \textit{top-5}, and \textit{top-10} accuracy for evaluating the recommendation system. The experiment is repeated $5$ times and the average of the \textit{top-k} accuracy are reported in Table~\ref{tab:table1}. In the first experiment, we only use the demographic information of users in the attention mechanism of the proposed recommendation system. Table~\ref{tab:table1} \textit{row 1} shows the accuracy. In the second, experiment, we used all information extracted from questionnaires in addition to their demographic information for the attention mechanism, and observed a slight decline in the accuracy (Table~\ref{tab:table1} \textit{row 2}). We hypothesize that most people don't have an accurate evaluation of their own ability. Thus, the answers to the questionnaires may not accurately represent their profiles, leading to a conflict to what exercises they performed and what they answered. For example, two persons may give exact similar answers to the questions (possibly not accurate answers), but have different ability and interest to perform exercises. In the rest of this paper, we only use demographic information to represent the user's profile.

We compared the baseline model with user demographic information with a slightly different network architecture to study the affect of the architecture. In the new architecture, we reduced the dimension of the RNN module to half, and also we reduce the dimension of the output module. The top-1, top-5, and top-10 accuracy are $60.87\%$, $90.55\%$, and $94.71\%$. Compared to the row 1 of Table~\ref{tab:table1}, the alternative architecture has a worse performance. The original structure proposed in Section~\ref{sec:network-arch} has the best performance for the mHealth data and is selected empirically.

\vspace{2mm}

\noindent \textbf{Baseline with Data Augmentation:} The training dataset was augmented with two approaches described in \S~\ref{sec:augment}. The training and augmented data were used to train the baseline model and evaluated as described in the baseline section. Table~\ref{tab:table1} \textit{row 3} shows the accuracy results of the baseline model trained with the training data and the expert augmented data. The data augmentation generalizes the model and improves the accuracy compared to the baseline (Table~\ref{tab:table1} \textit{row 1}).

On the other hand, we see a decline in the accuracy of the baseline model when it is trained on the training dataset and the augmented data by association rule mining algorithms (Table~\ref{tab:table1} \textit{row 4}). This observation points to the importance of having the expert in the loop for exercise recommendation systems. Because in this experiment the augmented dataset with expert knowledge gives higher accuracy, we use this method for other experiments in this paper. 

\vspace{2mm}

\noindent \textbf{Baseline with Active Learning:} The training data was used to estimate the parameters of the Dirichlet distributions: $[y_{(N)}, y_{(N-1)}, y]^T \sim \mathcal{D}(1.59, 0.42, 0.31)$. The \textit{marginal distribution} is calculated numerically, and the $\alpha=0.01$-level hypothesis testing results in:

\begin{equation*}
    \begin{cases}
      H_0: & z^i(t) \geq 0.18 \\
      H_1: & z^i(t) < 0.18
    \end{cases}
\end{equation*}

\noindent where $P(z^i(t) \geq 0.18)>1-\alpha$. During the test, if $z^i(t)$ falls bellow $0.18$, then the recommender asks the expert for the feedback and fine-tunes the network with the feedback. Because we are evaluating the proposed model on a dataset collected in the past, we cannot ask for the expert for feedback in our evaluation. Therefore, whenever $z^i(t)$ falls bellow $0.18$, we take the actual exercise the test participant performed at that time step and provide it as the feedback by an expert to our active learning algorithm (Table~\ref{tab:table1} \textit{row 5}). Comparing the results with our baseline model, the top-$1$ accuracy is increased by $10\%$. The increased in accuracy is the results of fine-tuning the recommender system with the feedback received from the expert, which makes the recommendation system personalized.

\vspace{2mm}

\noindent \textbf{Baseline with New User Initialization:} We calculated the pairwise similarity across all participants. For each new participant, we fine-tuned the recommendation system with the training data of the three most similar participants based on their profiles. The recommendation system is initialized for the new user by the fine-tuned network. We didn't use data augmentation and active learning in this experiment just to examine the effect of new user initialization. We see a minor improvement in top-$1$ accuracy (Table~\ref{tab:table1} \textit{row 6}). However, we believe that when it comes to more diverse participants, e.g., different age groups, race, ethnicity, the proposed initialization strategy improve the accuracy significantly.

\vspace{2mm}

\noindent \textbf{Baseline with Data Augmentation and Active Learning:} In this experiment, we combined the data augmentation and active learning as we hypothesise that the accuracy will improve. Table~\ref{tab:table1} \textit{row 7} shows the results. As we expected, the accuracy has improved with respect to only active learning (Table~\ref{tab:table1} \textit{row 5}) and only augmentation (Table~\ref{tab:table1} \textit{row 3}). 

\vspace{2mm}

\noindent \textbf{Baseline with Data Augmentation, New User Initialization, and Active Learning:} In the last experiment, we combined all modules. Table~\ref{tab:table1} \textit{row 9} represents the results indicating a very minor decline in accuracy by adding the new user initialization procedure (compared to Table~\ref{tab:table1} \textit{row 7}). More study with different questionnaires, medical records, etc. may lead to increase in the accuracy of the new user initialization method.

\setlength{\arrayrulewidth}{1.5pt}
\begin{table}\caption{The accuracy of the recommendation system.} \label{tab:table1}
 \begin{tabular} {| >{\arraybackslash}m{3.2cm}| >{\centering\arraybackslash}m{1.2cm}| >{\centering\arraybackslash}m{1.2cm}| >{\centering\arraybackslash}m{1.2cm}|}  %{||c{1em} c{1em} c{1em}||} %
 \hline
 \multicolumn{1}{|c|}{Method} & top-1 Accuracy & top-5 Accuracy & top-10 Accuracy\\ [0.5ex] 
 \hline\hline
 {\bf 1.} Baseline (Demographic) & 63.78\% &92.19\% &97.33\% \\ 
 \hline
 {\bf 2.} Baseline (Full Profile) &  61.16\% & 90.76\% & 96.98\% \\ 
 \hline
 {\bf{3.}} Baseline + Data Augmentation (Expert) & 72.53\% & 95.53\%& 98.56\%\\
 \hline
 {\bf{4.}} Baseline + Data Augmentation (Rule based) & 69.74\%  & 95.28\% & 98.68\%\\
 \hline
 {\bf{5.}} Baseline (Demographic) + Active Learning & 74.45\% & 95.29\% & 98.48\%\\
 \hline
 {\bf{6.}} Baseline (Demographic) + New user Init & 65.91\% & 93.14\% & 97.65\% \\ 
 \hline
%  {\bf{7.}} Baseline (Demographic) + New user Init + Active Learning & 75.79\% & 95.75\% & 98.58\% \\ 
%  \hline
%  {\bf{7.}} Baseline + Data Augmentation (Rule based) + Active Learning & 71.96???\% & 95.48???\% & 98.70???\%\\
 %\hline
 {\bf{7.}} Baseline + Data Augmentation (Expert) + Active Learning & {\bf{80.12\%}} & {\bf{97.23\%}} & {\bf{99.26\%}}\\
 \hline
 {\bf{8.}} Baseline + Data Augmentation (Expert) + New user Init & 71.90\% & 95.27\% & 98.42\%\\
 \hline
 {\bf{9.}} Baseline + Data Augmentation (Expert) + New user Init + Active Learning & 80.08\% & 97.00\% & 99.11\%\\
 %\hline
% {\bf{10.}} Baseline + Data Augmentation (Rule based) + New user Init & 63.45\% & 91.86\% & 97.27\%\\
% \hline
%  {\bf{10.}} Baseline + Data Augmentation (Rule based) + New user Init + Active Learning & 78.80\% & 95.72\% & 98.42\%\\

 \hline
\end{tabular}
\end{table}

\vspace{2mm}

\noindent \textbf{Comparison with the state-of-the-art:} The proposed model is evaluated against some of the well-known sequential recommendation systems \cite{hidasi2015session, covington2016deep, kalchbrenner2016neural, kula2017mixture} $^\dagger$ \footnote{$^ \dagger$ Implemented in \cite{kula2017spotlight}.}. Because the sequential models, including the proposed method, use the history of items for each users to propose the next item, they cannot be compared directly with matrix factorization approaches, e.g. SVD. The factorization approaches estimate the rating of different items without explicitly recommending the next item. However, the sequential models take into account the order of items in the history and recommend the next item without estimating the user's rating. 

To have a fair comparison, the baseline model with user demographic is used in this experiment. Table~\ref{tab:table2} shows the Top-\textit{k} accuracy for different models. The proposed baseline model outperforms the existing state-of-the-art in \textit{top-1}, \textit{top-5}, \textit{top-10} accuracy. The best performance after the proposed baseline method in Table~\ref{tab:table2} belongs to the CNN method \cite{kalchbrenner2016neural}. However, the performance of the CNN method is not as good as the proposed method. The superior performance of the proposed baseline model extends to the other variations of the proposed model, \textit{e.g.} the baseline with active learning.

\setlength{\arrayrulewidth}{1.5pt}
\begin{table}\caption{Comparison of the accuracy across different recommendation systems for the mHealth data.} \label{tab:table2}
 \begin{tabular} {| >{\arraybackslash}m{3.2cm}| >{\centering\arraybackslash}m{1.2cm}| >{\centering\arraybackslash}m{1.2cm}| >{\centering\arraybackslash}m{1.2cm}|}  %{||c{1em} c{1em} c{1em}||} %
 \hline
 \multicolumn{1}{|c|}{Method} & top-1 Accuracy & top-5 Accuracy & top-10 Accuracy\\ [0.5ex] 
 \hline\hline
 GRU4REC & 7.89\% &22.74\% &40.04\% \\ 
 \hline
 Pooling &  8.83\% & 41.16\% & 50.38\% \\ 
 \hline
 CNN & 12.97\% & 33.83\%& 53.00\%\\
 \hline
 Mixture & 5.45\%  & 20.48\% & 40.41\%\\
 \hline
 Baseline (Demographic) & \bf{63.78}\% &\bf{92.19}\% &\bf{97.33}\% \\
 \hline
\end{tabular}
\end{table}

\subsection{MovieLens}

In the second experiment, the proposed method is applied to the MovieLens100K dataset \cite{harper2015movielens} and compared with the state of the art sequential recommendation systems \cite{hidasi2015session, covington2016deep, kalchbrenner2016neural, kula2017mixture}. The dataset contains 100,000 user IDs, movie IDs, the user ratings of movies, and timestamps. We group the dataset by the user IDs and sorted them with their timestamps. %MovieLens is a dense dataset \cite{ji2020sequential}

\subsubsection{\textbf{Experiment Setup and Discussion}} 

Our proposed recommendation system was developed specifically for recommending exercises, when the number of exercises is limited (unlike the number of movies). However, the proposed method is evaluated on the MovieLens dataset to show its strength in scaling to other applications. To this end, we pre-processed the dataset and limited the number of movies in this dataset to the 100 most watched movies in Movielens100K. Thus, the new subset of MovieLens100K is more comparable to the exercises activity dataset when the number of items are limited.  

Similar to the mHealth dataset, we selected the window length of $w=3$ resulted in $29931$ sequence samples, and are split to 80\% training and 20\% test samples. The architecture of the proposed recommendation system is selected empirically as follows: $K_X=K_u=1000$, $K_e=3$, and $K=300$. Table~\ref{tab:table3} shows \textit{top-1}, \textit{top-5}, and \textit{top-10} accuracy of different recommendation systems.  In recommending new movies in the subset of MovieLens100K, the proposed method surpasses the accuracy of the state of the art methods. After the proposed method, the pooling method \cite{covington2016deep} comes second with accuracy less than the proposed baseline method.

\setlength{\arrayrulewidth}{1.5pt}
\begin{table}\caption{Comparison of the accuracy across different recommendation systems for a subset of MovieLens100K.} \label{tab:table3}
 \begin{tabular} {| >{\arraybackslash}m{3.2cm}| >{\centering\arraybackslash}m{1.2cm}| >{\centering\arraybackslash}m{1.2cm}| >{\centering\arraybackslash}m{1.2cm}|}  %{||c{1em} c{1em} c{1em}||} %
 \hline
 \multicolumn{1}{|c|}{Method} & top-1 Accuracy & top-5 Accuracy & top-10 Accuracy\\ [0.5ex] 
 \hline\hline
 GRU4REC & 2.5\% &12.27\% &21.55\% \\ 
 \hline
 Pooling & 3.93\% & 18.91\% & 39.66\% \\ 
 \hline
 CNN & 2.26\% & 10.91\%& 20.23\%\\
 \hline
 Mixture & 2.47\%  & 11.37\% & 21.34\%\\
 \hline
 Baseline (Demographic) & \bf{5.90}\% &\bf{26.38}\% &\bf{46.32}\% \\
 \hline
\end{tabular}
\end{table}

%The experiment is also conducted on the full Movielens100K dataset. As expected, the accuracy generally drops because there are much more items (movies) for the recommendation system to learn from. However, our proposed approach still outperforms the state-of-the-art. 

%\setlength{\arrayrulewidth}{1.5pt}
%\begin{table}\caption{Comparison of the accuracy across different recommendation systems for MovieLens100K.} \label{tab:table4}
% \begin{tabular} {| >{\arraybackslash}m{3.2cm}| >{\centering\arraybackslash}m{1.2cm}| >{\centering\arraybackslash}m{1.2cm}| >{\centering\arraybackslash}m{1.2cm}|}  %{||c{1em} c{1em} c{1em}||} %
% \hline
% \multicolumn{1}{|c|}{Method} & top-1 Accuracy & top-5 Accuracy & top-10 Accuracy\\ [0.5ex] 
% \hline\hline
% GRU4REC & 0.38\% &2.30\% &4.38\% \\ 
% \hline
% Pooling & 1.22\% & 5.60\% & 10.27\% \\ 
% \hline
% CNN & 0.31\% & 2.21\%& 4.20\%\\
% \hline
% Mixture & 0.56\%  & 2.00\% & 3.67\%\\
% \hline
% Baseline (Demographic) & \bf{5.90}\% &\bf{26.38}\% &\bf{46.32}\% \\
% \hline
%\end{tabular}
%\end{table}

\section{Conclusion}

In this paper, a physical exercise recommendation system was developed. The main challenge in developing recommendation systems is to make them personalized, especially for new users when the training dataset does not exist. Analyzing the outcomes of the experimental results indicates the importance of user and exercise profiles as attention mechanisms.Thus, the developed system took advantage of the health and demographic questionnaires filled out by users before joining the program to use as attention mechanism.

In spite of all fine tuning and the attention mechanism, the experimental results show that the perfect personalization cannot be achieved. While constant feedback from the user interaction can be collected in e-commerce, music and movie recommendation systems, having an expert in the loop is inevitable in exercise recommendation systems because we don't know whether the user performed the exercise or not (while in e-commerce the user either buys the product or listen completely to a music on a music platform). The expert knowledge provides information when the recommender is uncertain and fails to make accurate predictions, which is why a real time active learning mechanism in conjunction with our developed recommendation systems showed a significant increase in the accuracy of the system. 

\begin{figure}[!tbp]%\label{fig:appendix}

  \centering
  \subfloat[]{\includegraphics[width=0.22\textwidth]{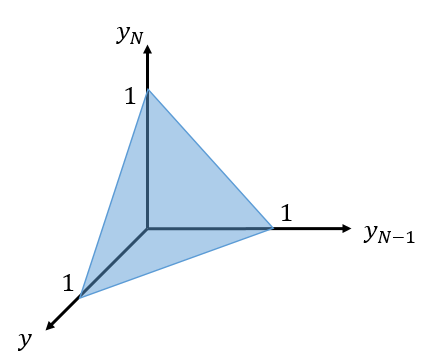}\label{fig:f1}}
  \hfill
  \subfloat[]{\includegraphics[width=0.24\textwidth]{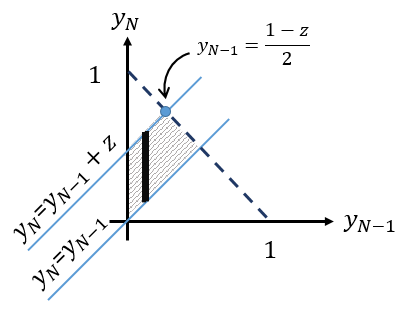}\label{fig:f2}}
  \caption{(a) The valid surface of $y_{(N)}, y_{(N-1)}, y$ variables with the Dirichlet distribution. (b) The integral area projected on $y_{(N)}-y_{(N-1)}$ plane.}
\end{figure} 

\section*{Acknowledgment}

The authors would like to thank Dr. Choh Man Teng for assisting with the expert studies.

% %Dr. Reveryrand would like to acknowledge the funding by XLIM, Limoges, France. 
% The authors would like to thank Dr. David Root and Dr. Jean-Pierre Teyssier at Agilent Technologies for the loan of the time-domain nonlinear measurement equipment and TriQuint Semiconductor for the donation of the transistors. 

%*********************************************************************
%*********************************************************************
\begin{figure*}[!t]
% ensure that we have normalsize text
\normalsize
% Store the current equation number.
\setcounter{MYtempeqncnt}{\value{equation}}
% Set the equation number to one less than the one
% desired for the first equation here.
% The value here will have to changed if equations
% are added or removed prior to the place these
% equations are referenced in the main text.
\setcounter{equation}{10}
\begin{multline}
\label{eq:11}
P(y_{(N)}-y_{(N-1)} \leq z)= \int_{y_{(N-1)}=0}^{1-z \over 2}{\int_{y_{(N)}=y_{(N-1)}}^{y_{(N-1)}+z} {f(y_{(N)},y_{(N-1)},1-y_{(N)}-y_{(N-1)})}} dy_{(N)} dy_{(N-1)} \\
    +\int_{y_{(N-1)}={1-z \over 2}}^{0.5}{\int_{y_{(N)}=y_{(N-1)}}^{y_{(N-1)}+z}{f(y_{(N)},y_{(N-1)},1-y_{(N)}-y_{(N-1)})}} dy_{(N)} dy_{(N-1)}
\end{multline}
\begin{multline}
\label{eq:12}
f(z)= {-1 \over 2}\int_{y_{(N)}={1-z \over 2}}^{{1+z} \over 2}{f(y_{(N)},{1-z \over 2},1-y_{(N)}-{1-z \over 2})} dy_{(N)} \\
    +\int_{y_{(N-1)}=0}^{1-z \over 2} {f(y_{(N-1)}+z,y_{(N-1)},1-2y_{(N-1)}-z)} dy_{(N-1)}
    +{1 \over 2}\int_{y_{(N)}={1-z \over 2}}^{{1+z \over 2}}{f(y_{(N)},{1-z \over 2},1-y_{(N)}-{1-z \over 2})} dy_{(N)} \\
    =\int_{y_{(N-1)}=0}^{1-z \over 2} {f(y_{(N-1)}+z,y_{(N-1)},1-2y_{(N-1)}-z)} dy_{(N-1)}
\end{multline}
\begin{equation}
\label{eq:13}
    f(z)= {  {\Gamma (\alpha_1+\alpha_2+\alpha_3)} \over {\Gamma(\alpha_1)\Gamma(\alpha_2)\Gamma(\alpha_3)}} 
     \int_{y_{(N-1)}=0}^{1-z \over 2}{
     {(y_{(N-1)}+z)}^{(\alpha_1-1)} y_{(N-1)}^{(\alpha_2-1)} {(1-2y_{(N-1)}-z)}^{(\alpha_3-1)} dy_{(N-1)}}
\end{equation}

% Restore the current equation number.
\setcounter{equation}{\value{MYtempeqncnt}}
% The IEEE uses as a separator
\hrulefill
% The spacer can be tweaked to stop underfull vboxes.
\vspace*{4pt}
\end{figure*}
%*********************************************************************
%*********************************************************************

% if have a single appendix:
\appendices
%\appendix[Probability Distribution of the \textit{Marginal Distance}]
\section{Probability Distribution of the \textit{Marginal Distance}}
\label{sec:appendix}

In \textit{marginal distance}, the output variables are sorted in ascending order as $y_{(1)} \leq y_{(2)} \leq \ldots \leq y_{(N)}$. The \textit{marginal distance} is defined as $M=y_{(N)}-y_{(N-1)}$. When the value of the \textit{marginal distance} falls below a given threshold $\theta_m$, the recommender asks the human expert for labeling. While in prior works the threshold was determined by users, we define the probability distribution of the \textit{marginal distance}. The output variable is drawn from a Dirichlet probability distribution function, $\mathcal{D}(\alpha_1, \alpha_2, \ldots, \alpha_N)$. In the Dirichlet distribution, the amount of the variables sum up to one: $\sum_{k=1}^N{y_{(k)}}=1$. In the \textit{marginal distance}, we are only interested in $y_{(N)}$ and $y_{(N-1)}$. To simplify the calculation, we define a new random variable $y=\sum_{k=1}^{N-2}{y_{(k)}}$. Note that the marginal distribution of $y_{(k}$s are Beta distribution and the summation of several Beta distribution, $y$, is a Beta distribution. The joint distribution of $Y=[y_{(N)}, y_{(N-1)}, y]^T$ is a Dirichlet distribution, $\mathcal{D}(\alpha_1, \alpha_2, \alpha_3)$,:

\begin{equation}
    f(y_{(N)}, y_{(N-1)}, y)  = {\Gamma (\sum_k{\alpha_k}) \over \prod_k{\Gamma(\alpha_k)}} y_{(N)}^{(\alpha_1-1)} y_{(N-1)}^{(\alpha_2-1)} y^{(\alpha_3-1)}
\end{equation}

\noindent where $0 \leq y_{(N)}, y_{(N-1)}, y <1$, and $y_{(N)}+y_{(N-1)}+y=1$. The parameters $\alpha_1, \alpha_2$ and $\alpha_3$ are estimated from the training dataset using Maximum Likelihood approach \cite{minka2000estimating, ericsuh}.  

The probability distribution of the \textit{marginal distance} is derived from transforming $y_{(N)}$ and $y_{(N-1)}$ based on $Z=y_{(N)}-y_{(N-1)}$. Because the support of $y_{(N)}, y_{(N-1)}, y$ is restricted to the hyper-plane defined by $y_{(N)}+y_{(N-1)}+y=1$, the third argument, $y$, is known given $y_{(N)}$ and $y_{(N-1)}$. Fig.~\ref{fig:f1} shows the $y_{(N)}+y_{(N-1)}+y=1$ hyper-plane. Because the support of $y_{(N)}, y_{(N-1)}, y$ is restricted and the value of $y$ depends on $y_{(N)}$ and $y_{(N-1)}$, we can just visualize the projection of the hyperplane on the $y_{(N)}-y_{(N-1)}$ plane. This makes defining the boundaries of the integrals easier. Fig.~\ref{fig:f2} shows this projection with $y_{(N)} > y_{(N-1)}$ and $Z=y_{(N)}-y_{(N-1)}$ lines. In the first step, we derive the cumulative probability distribution $F(z)=P(Z \leq z)$, as in Eq.~\ref{eq:11}. Then, to get the probability distribution function, we take the derivatives from both sides with respect to $z$ to obtain Eq.~\ref{eq:12}. Thus, we get the probability distribution function the \textit{marginal distance} as in Eq.~\ref{eq:13}. We calculate $f(z)$ numerically by approximating the integral with a summation for different value of $z$, even though calculation of the closed form is not impossible.

% \begin{multline}
%     \label{eq:11}
%     P(y_{(N)}-y_{(N-1)} \leq z)= \\ \int_{y_{(N-1)}=0}^{1-z \over 2}{\int_{y_{(N)}=y_{(N-1)}}^{y_{(N-1)}+z} {f(y_{(N)},y_{(N-1)},1-y_{(N)}-y_{(N-1)})}} dy_{(N)} dy_{(N-1)}\\
%     +\int_{y_{(N-1)}={1-z \over 2}}^{0.5}{\int_{y_{(N)}=y_{(N-1)}}^{y_{(N-1)}+z}{f(y_{(N)},y_{(N-1)},1-y_{(N)}-y_{(N-1)})}} dy_{(N)} dy_{(N-1)}
% \end{multline}
  
% To get the probability distribution function, we take the derivatives from both sides with respect to $z$:

% \begin{multline}
% \label{eq:12}
%     f(z)\\= {-1 \over 2}\int_{y_{(N)}={1-z \over 2}}^{{1+z} \over 2}{f(y_{(N)},{1-z \over 2},1-y_{(N)}-{1-z \over 2})} dy_{(N)}\\
%     +\int_{y_{(N-1)}=0}^{1-z \over 2} \\ {f(y_{(N-1)}+z,y_{(N-1)},1-2*y_{(N-1)}-z)} dy_{(N-1)}\\
%     {1 \over 2}\int_{y_{(N)}={1-z \over 2}}^{{1+z \over 2}}{f(y_{(N)},{1-z \over 2},1-y_{(N)}-{1-z \over 2})} dy_{(N)}\\
%     =\int_{y_{(N-1)}=0}^{1-z \over 2} \\{f(y_{(N-1)}+z,y_{(N-1)},1-2*y_{(N-1)}-z)} dy_{(N-1)}
% \end{multline}

% Thus, the probability distribution function the \textit{marginal distance} is:

% \begin{multline}
% \label{eq:13}
%     f(z)= {  {\Gamma (\alpha_1+\alpha_2+\alpha_3)} \over {\Gamma(\alpha_1)\Gamma(\alpha_2)\Gamma(\alpha_3)}} \\
%      \int_{y_{(N-1)}=0}^{1-z \over 2}{
%      {(y_{(N-1)}+z)}^{(\alpha_1-1)} y_{(N-1)}^{(\alpha_2-1)}\\ {(1-2y_{(N-1)}-z)}^{(\alpha_3-1)} dy_{(N-1)}}
% \end{multline}

% We calculate $f(z)$ numerically by approximating the integral with a summation for different value of $z$, even though calculation of the closed form is not impossible. 

\hfill $\boxempty$

\section{Computational Cost}
\label{sec:appendix2}

In this section ,the computational cost of different modules are calculated. The computational cost of the exercise input module is $\mathcal{O}(N)$ because $N>K_x$. The computational cost of the User Profile input module is $\mathcal{O}(N_U)$, and that of the Exercise Profile input is $\mathcal{O}(N_e)$. The computational cost of the RNN module is $\mathcal{O}(K^\omega)$. $K$ is smaller than $N$ but not a lot smaller. So the overall computational cost is dominated by the RNN module. Thus, the overal computational cost is $\mathcal{O}(K^\omega)$.

\hfill $\boxempty$

% or
%\appendix  % for no appendix heading
% do not use \section anymore after \appendix, only \section*
% is possibly needed

% use appendices with more than one appendix
% then use \section to start each appendix
% you must declare a \section before using any
% \subsection or using \label (\appendices by itself
% starts a section numbered zero.)
%

% ============================================
%\appendices
%\section{Proof of the First Zonklar Equation}
%Appendix one text goes here %\cite{Roberg2010}.

% you can choose not to have a title for an appendix
% if you want by leaving the argument blank
%\section{}
%Appendix two text goes here.

% use section* for acknowledgement
%\section*{Acknowledgment}

%The authors would like to thank D. Root for the loan of the SWAP. The SWAP that can ONLY be usefull in Boulder...

% Can use something like this to put references on a page
% by themselves when using endfloat and the captionsoff option.
\ifCLASSOPTIONcaptionsoff
  \newpage
\fi

% trigger a \newpage just before the given reference
% number - used to balance the columns on the last page
% adjust value as needed - may need to be readjusted if
% the document is modified later
%\IEEEtriggeratref{8}
% The "triggered" command can be changed if desired:
%\IEEEtriggercmd{\enlargethispage{-5in}}

% ====== REFERENCE SECTION

%\begin{thebibliography}{1}

%\newpage
% IEEEabrv,

\bibliographystyle{IEEEtran}
\bibliography{IEEEabrv,Bibliography}

\vfill

% Can be used to pull up biographies so that the bottom of the last one
% is flush with the other column.
%\enlargethispage{-5in}

% that's all folks
\end{document}